\documentclass[10pt,twocolumn,letterpaper]{article}
\usepackage{vivid}
\usepackage{xcolor}
\usepackage{tcolorbox}
\tcbuselibrary{skins,breakable}
\newtcolorbox[auto counter, number within=section]{listingbox}[2][]{%
  enhanced,
  breakable,
  colback=gray!5,
  colframe=black!30,
  coltitle=black,
  fonttitle=\bfseries,
  title=Listing~\thetcbcounter:~#2,
  sharp corners,
  boxrule=0.8pt,
  left=6pt, right=6pt, top=6pt, bottom=6pt,
  before skip=10pt, after skip=10pt,
  #1
}
\usepackage{enumitem}
\usepackage{graphicx}
\usepackage{booktabs,multirow,tabularx,array}
\usepackage{url}
\usepackage{microtype}
\usepackage{iftex}
\definecolor{linkblue}{rgb}{0.21,0.49,0.74}
\usepackage[pagebackref,breaklinks,colorlinks,allcolors=linkblue]{hyperref}
\title{VIVID: A Visual Harness for Agentic Video Editing}
\author{%
  Wenhan Yu$^{1}$ \quad Zhaoxi Zhang$^{2}$ \quad Zhilong Tan$^{1}$ \\
  Weikang Li$^{2,\dagger}$ \quad Lei Sha$^{1,\dagger}$ \\[4pt]
  \normalsize $^{1}$Beihang University \quad $^{2}$Peking University \\
  \small\texttt{yuwenhan@buaa.edu.cn} \quad \texttt{zhaoxizhang25@stu.pku.edu.cn} \quad \texttt{rongbit@sina.com} \\
  \small\texttt{wavejkd@pku.edu.cn} \quad \texttt{shalei@buaa.edu.cn} \\[2pt]
  \small $^{\dagger}$Corresponding authors.
}
\begin{document}
\maketitle
\begin{abstract}

Video editing requires agents to make editing decisions grounded in audiovisual observations and precise temporal localization, and to evaluate their effects in rendered outputs. General-purpose coding agents possess strong comprehension, reasoning, and code execution capabilities, while media interaction and evidence management mechanisms tailored to video tasks can further unlock their editing potential. We introduce \textbf{VIVID}, a \textbf{vi}sual harness that supports multiple native coding agents in \textbf{vid}eo post-production without additional model training. VIVID provides tools for on-demand video inspection, audio analysis, and transcription, and preserves observations alongside their corresponding media versions and temporal coordinates. Agents can actively retrieve these observations as they reason, perform edits using their native coding and shell tools, and inspect rendered outputs to guide subsequent revisions. On AgenticVBench, comprising 100 tasks across Assembly, Repair, Sequencing, and Repurpose, Codex with VIVID achieves an average score of 58.6\%, an improvement of 20.2 percentage points over native Codex. The same harness also integrates with Claude Code and OpenCode, achieving average scores of 55.1\% and 37.1\%, respectively, both substantially exceeding their native baselines. These results highlight VIVID's potential as a reusable visual harness for extending the capabilities of general-purpose coding agents to video editing.
\end{abstract}

\section{Introduction}
\label{sec:introduction}

Video editing requires editors to understand source materials, select appropriate content, and transform it into a work that meets the task requirements through cutting, sequencing, and audiovisual adjustments~\cite{leake2017computational,huber2019b,wang2024lave}. This process calls for continual observation and judgment: whether a shot fits the narrative, where a line of dialogue should end, and whether captions align with the image and sound all depend on the underlying audiovisual content. Editors must also inspect the rendered video, as a well-reasoned editing plan does not always translate into a correct output. They therefore need to review the result and revise it when necessary.

\begin{figure}
    \centering
    \includegraphics[width=\linewidth]{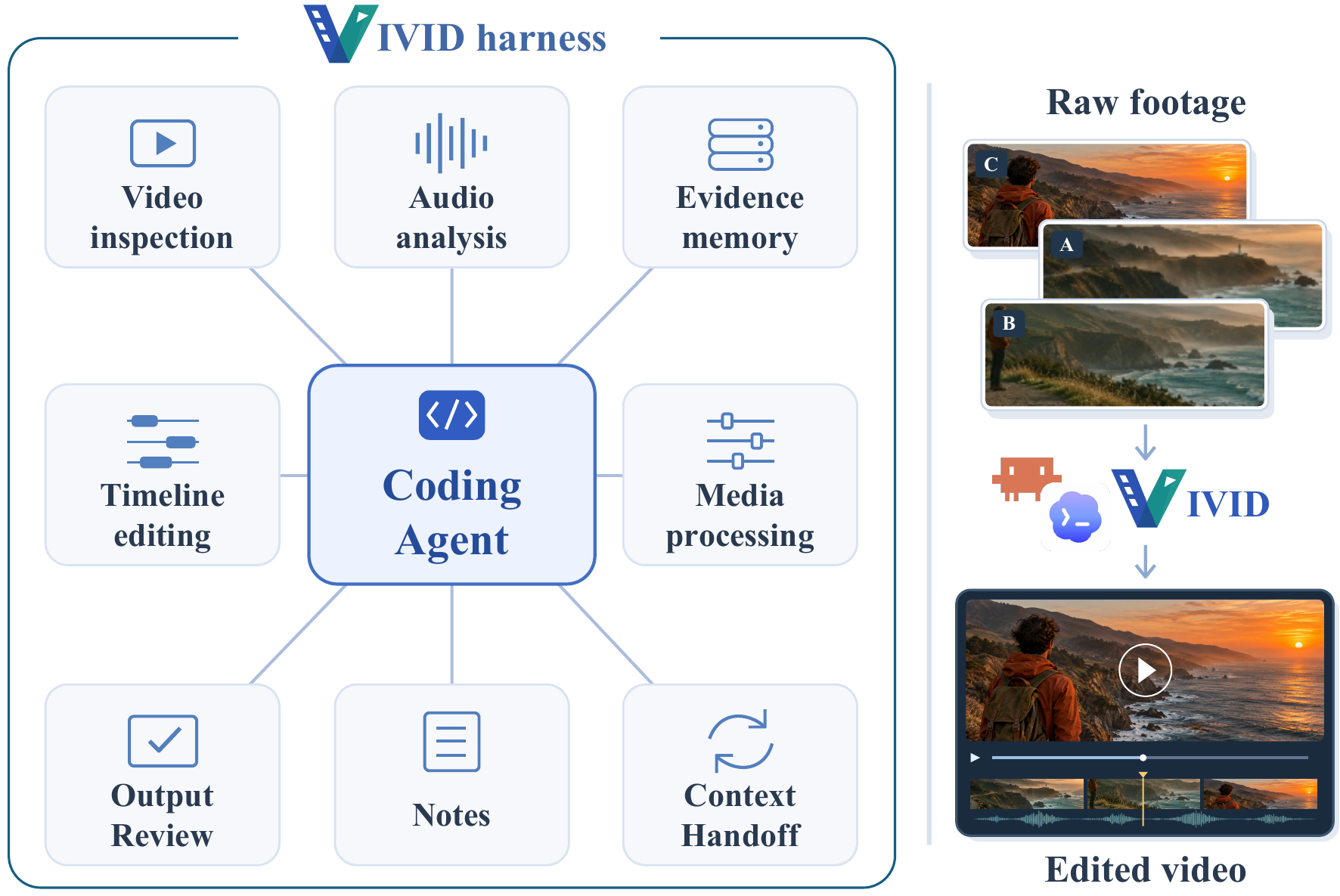}
    \caption{Overview of VIVID, a visual harness for agentic video editing. VIVID augments native harnesses with on-demand audiovisual inspection, persistent evidence memory and retrieval, and output inspection and revision to support iterative editing from source footage to rendered video.}
    \label{fig:teaser}
\end{figure}

General-purpose coding agents possess language understanding, reasoning, and code execution capabilities. With support from an external harness, they can invoke media tools to perform operations such as cutting, concatenation, and captioning, providing a foundation for automated video editing~\cite{openai2025codex,anthropic2025claudecode,wang2024openhands,zhang2025one,gao2025trae}. Prior work has explored specialized editing workflows, role collaboration, and the management of intermediate artifacts to support video production~\cite{sandoval2025editduet,yan2026crayotter}. AgenticVBench further evaluates agents on real-world post-production tasks~\cite{cao2026agenticvbench}. These advances raise a question: how can appropriate media interaction interfaces further unlock the video editing capabilities of general-purpose coding agents while preserving their native tools and execution environments?

A central challenge is that the evidence needed for editing changes as reasoning and editing progress. An initial overview may reveal candidate shots but provide insufficient detail to determine precise cut boundaries. Likewise, an earlier description of the content may omit details needed to adjust captions or repair audiovisual synchronization. Research on active video understanding and visual memory suggests that agents can select observations according to their current questions and revisit existing evidence when needed~\cite{fan2024videoagent,liu2024video,zhou2026videoagent,wang2025videotree,wang2025videodirector}. Video editing further requires this evidence to be associated with specific media versions and temporal positions: cutting, reordering, and speed changes alter where content appears in the output, while successive renders produce different versions. Agents therefore need traceable connections among source observations, editing decisions, and output inspection.

As shown in Fig.~\ref{fig:teaser}, we introduce \textbf{VIVID}, a visual harness for agentic video editing. VIVID provides native coding agents with on-demand audiovisual observation, persistent evidence, and output inspection. These capabilities allow agents to locate relevant content, retrieve observations associated with media versions and temporal coordinates, and revise edits based on rendered results. Agents can choose their own observation and editing strategies, using either native tools or VIVID's optional editing interface, without additional model training or updates. On the 100 tasks of AgenticVBench, Codex, Claude Code, and OpenCode with VIVID achieve average scores of 58.6\%, 55.1\%, and 37.1\%, improving upon their respective native configurations by 20.2, 22.7, and 9.7 percentage points. Ablation studies and trajectory analysis further support the value of media tools and evidence management, and show improvements in content selection, temporal localization, and consistency between editing plans and outputs. These results demonstrate that VIVID can extend the video editing capabilities of general-purpose coding agents.

Our contributions are summarized as follows:
\begin{itemize}
    \item We introduce \textbf{VIVID}, which integrates on-demand audiovisual observation, persistent evidence associated with media versions and temporal coordinates, and output inspection into the video editing workflows of native coding agents.
    
    \item We demonstrate the effectiveness of VIVID on AgenticVBench, improving editing performance across three model--harness configurations without additional model training.
    
    \item We analyze VIVID's performance gains, execution behavior, and computational overhead through ablation studies, tool-call and token-consumption analyses, editing trajectories, and case studies.
\end{itemize}

\section{Related Work}
\label{sec:related}

\subsection{Agentic video editing}
Computational video editing combines source material understanding with editorial decision-making, spanning tasks such as dialogue-scene assembly, transcript-based editing, and language-guided timeline manipulation~\cite{leake2017computational,huber2019b,wang2024lave,tilekbay2024expressedit}. Recent work further explores how specialized models and agentic workflows can follow natural-language instructions to produce coherent video edits~\cite{pardo2024generative,sandoval2025editduet,ding2025prompt,zhou2026videoagent}.

For example, EditDuet coordinates an Editor and a Critic for nonlinear editing, while VideoAgent combines shot planning with specialized editing agents~\cite{sandoval2025editduet,zhou2026videoagent}. Other systems address long-form video editing through semantic indexing, inspectable production artifacts, or learned editing policies~\cite{ding2025prompt,yan2026crayotter,yan2026crayotterlearning,mu2026weviscut}. AgenticVBench evaluates agents' ability to complete video post-production tasks through assignments involving Assembly, Repair, Sequencing, and Repurpose~\cite{cao2026agenticvbench}. VIVID supports video post-production by extending existing coding agents with media interaction capabilities, without updating the underlying model parameters, while allowing agents to choose how to observe and edit.

\subsection{Agent interfaces and editing harnesses}

The interface between an agent and its execution environment influences how the agent observes its surroundings, uses tools, and carries out multistep tasks~\cite{yang2024swe,wang2024executable,wang2025openhands,yao2022react,gupta2023visual}. In video editing, Crayotter preserves intermediate production artifacts and execution traces to support inspection and repair of the editing process~\cite{yan2026crayotter,yan2026crayotterlearning}. Other systems incorporate feedback and local revision into their editing workflows~\cite{zeng2026cassette,chen2026auto}. VIVID explores a visual harness that combines audiovisual inspection and evidence retrieval with the native tools and execution environments of coding agents. It supports source observation, output inspection, and verification, helping agents make more accurate editing decisions.

\subsection{Active video understanding}

Video understanding agents use selective observation and memory mechanisms to reason over long videos~\cite{wang2024videoagent}. The VideoAgent systems explore iterative frame retrieval and structured multimodal memory, respectively~\cite{fan2024videoagent}, while VideoTree and LensWalk adapt the temporal scope or granularity of observation to the current query~\cite{wang2025videotree,li2026lenswalk}. Other work investigates how to preserve visual observations or construct reusable memory for subsequent reasoning~\cite{han2026vista}. These works primarily focus on video understanding~\cite{suris2023vipergpt,li2026visual}. VIVID brings active observation and persistent evidence into a video editing workflow, allowing agents to select what to inspect based on the current task and use observations associated with media versions and temporal coordinates to guide subsequent edits and output inspection. This supports more precise video segment localization and editing decisions.

\begin{figure*}[t]
  \centering
  \includegraphics[width=\textwidth]{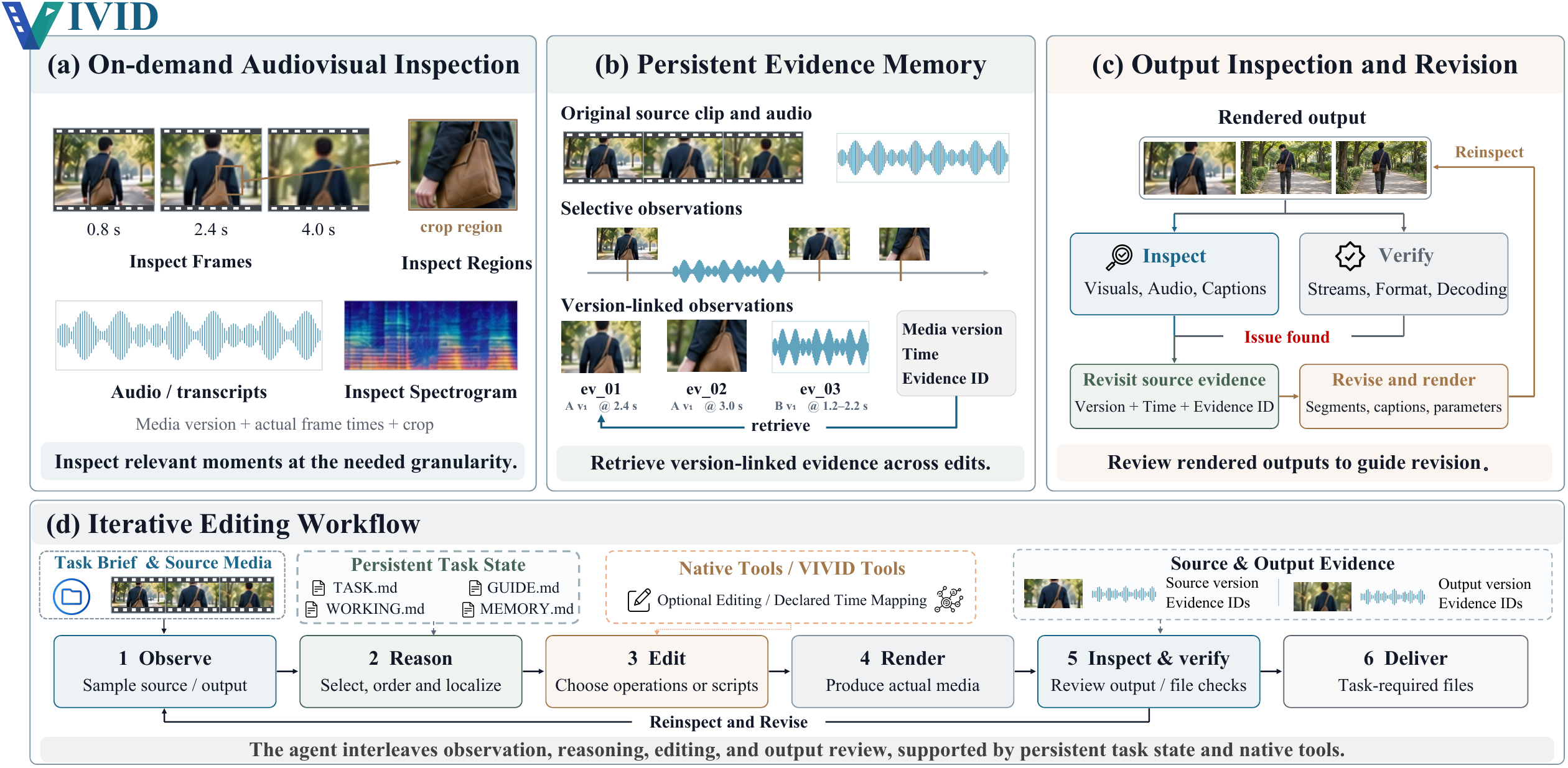}
  \caption{VIVID's core components and iterative editing workflow. The agent inspects media on demand, retains version-linked evidence for retrieval, and reviews rendered outputs to guide revisions.}
  \label{fig:method}
\end{figure*}

\section{Method}
\label{sec:method}

Video editing requires an agent to understand source materials, localize relevant segments, and inspect the actual output after editing. Because edits change the temporal positions of content and multiple rounds of editing can produce different media versions, the agent needs to revisit evidence associated with specific versions and time coordinates to support subsequent decisions and review.

To this end, we introduce VIVID, a set of extensions to native agent harnesses for video editing. Given a task brief and source materials, VIVID augments the native harness with on-demand audiovisual inspection, persistent evidence memory and retrieval, and output inspection and revision. These capabilities preserve the native harness's execution environment and the agent's control over its editing strategy. The agent decides what to inspect, how to perform edits, and when to review the output, while VIVID maintains traceable observations across the process. The method requires no additional training or updates to the underlying model.

\subsection{Core Components of VIVID}

\paragraph{On-demand audiovisual inspection.}
As shown in Fig.~\ref{fig:method}(a), VIVID allows the agent to select the temporal range and spatial region to inspect based on the current question. For video, the agent can first request a sparse overview across time, then sample frames or crop regions within candidate segments. It can also compare sampled frames from multiple source files, with filenames and timestamps, to support shot selection and ordering. Each observation includes images together with their media version, actual frame timestamps, and crop information, allowing visual judgments to be associated with specific positions in the source material.

Consider a visual inspection request specifying a media version $v$, a time range $I$, and an optional spatial region $R$. Let $T=\{t_1,\ldots,t_m\}$ be the timestamps of the frames actually returned. The observation can be represented as
\begin{equation}
    O_{\mathrm{vis}}(v,I,R)
    =
    \left\{
        \left(
            t_j,
            \operatorname{Crop}
            \bigl(\operatorname{Decode}(v,t_j),R\bigr)
        \right)
    \right\}_{j=1}^{m}.
\end{equation}
Here, $t_j$ is the time coordinate of the returned frame. When no region is specified, $\operatorname{Crop}$ retains the full frame.

For audio, VIVID can extract an audio window, provide amplitude statistics, waveforms, and spectrograms, and use speech recognition to produce segment-level text or estimated word-level timestamps. The agent can combine visual observations and transcripts to locate dialogue and then inspect potential cut points. This design allows the agent to choose the scope and granularity of inspection for the question and examine details as needed, supporting segment selection, temporal localization, and review of the final output.

\paragraph{Persistent evidence memory and retrieval.}
During editing, the agent may revisit the same segment or reconsider an earlier judgment after producing multiple versions. VIVID therefore stores acquired observations outside the model's current context, as shown in Fig.~\ref{fig:method}(b). The system assigns identifiers to media versions and associates each observation with its corresponding version, time coordinates, and evidence identifier. When the contents of a file change at the same path, existing evidence remains linked to the version from which it was obtained. This allows the agent to distinguish content across versions and revisit the observations underlying earlier decisions.

The agent can search indexed transcripts and other recorded text to locate relevant evidence, or revisit a saved observation through its evidence identifier. Version and time links support revisiting earlier judgments without confusing evidence from different media versions.

\paragraph{Output inspection and revision.}
After rendering, VIVID treats the actual output as an object of inspection, as illustrated in Fig.~\ref{fig:method}(c). The agent can review its visuals, audio, and captions, focusing on intervals around cuts or task-specific requirements. The agent combines format and decoding checks with audiovisual review of the output to guide revision. When an issue is found, the agent can revisit relevant source evidence, revise segments, captions, or processing parameters, and render and inspect a new output. This feedback connects editing decisions to their effects in the rendered media.

\subsection{Iterative Editing Workflow}

\begin{table*}[t]
\centering
\caption{ Performance of different model--harness configurations on AgenticVBench. All scores are percentages. Values reported with $\pm$ indicate the mean and standard deviation over three independent runs. Results from prior work are cited in the corresponding rows. ``--'' indicates that a result is unreported or not applicable. $\Delta$ and blue superscripts indicate percentage-point gains over the corresponding native harness for the overall average and each task category, respectively. Bold entries indicate VIVID-augmented configurations. }
\label{tab:main_results}

\resizebox{\textwidth}{!}{
\begin{tabular}{llcccccc}
\toprule
\textbf{Model} & \textbf{Harness} & \textbf{Average (\%)} & \textbf{$\Delta$} & \textbf{Repurpose} & \textbf{Sequencing} & \textbf{Repair} & \textbf{Assembly} \\
\midrule

Human Experts & -- & 88.5 & -- & 95 & 90 & 88 & 81 \\

\midrule

Qwen3.8-Omni-Flash & Claude Code~\cite{team2026qwen3} & 36.8 & -- & -- & -- & -- & -- \\

GPT-5.5 & Codex~\cite{cao2026agenticvbench} & $31.0 \pm 4.0$ & -- & 30 & 26 & 30 & 38 \\

Gemini 3.1 Pro & OpenCode~\cite{cao2026agenticvbench} & $23.8 \pm 3.7$ & -- & 23 & 19 & 20 & 33 \\

Crayotter-9B & Crayotter env.~\cite{yan2026crayotter} & 15.7 & -- & 23 & 7.9 & 15.2 & 14 \\

Qwen3.5-397B-A17B & Crayotter env.~\cite{yan2026crayotter} & 10.9 & -- & 15.3 & 8.1 & 9.6 & 7.5 \\

Seed 2.0 Lite & Claude Code~\cite{team2026qwen3} & 10.0 & -- & -- & -- & -- & -- \\

\midrule

\multirow{3}{*}{GPT-5.6-sol} & Crayotter env. & $33.7 \pm 2.9$ & -- & 30 & 34 & 23 & 53 \\

& Codex~\cite{cao2026agenticvbench} & $38.4 \pm 6.3$ & -- & 34 & 33 & 39 & 48 \\

& \textbf{Codex + VIVID} & $\mathbf{58.6 \pm 3.8}$ & \textbf{+20.2} & \textbf{58}$^{\textcolor{blue}{\uparrow 24}}$ & \textbf{58}$^{\textcolor{blue}{\uparrow 25}}$ & \textbf{43}$^{\textcolor{blue}{\uparrow 4}}$ & \textbf{77}$^{\textcolor{blue}{\uparrow 29}}$ \\

\addlinespace[3pt]

\multirow{2}{*}{Claude Fable 5} & Claude Code~\cite{cao2026agenticvbench} & $32.4 \pm 5.1$ & -- & 29 & 23 & 31 & 46 \\

& \textbf{Claude Code + VIVID} & $\mathbf{55.1 \pm 3.6}$ & \textbf{+22.7} & \textbf{56}$^{\textcolor{blue}{\uparrow 27}}$ & \textbf{59}$^{\textcolor{blue}{\uparrow 36}}$ & \textbf{37}$^{\textcolor{blue}{\uparrow 6}}$ & \textbf{66}$^{\textcolor{blue}{\uparrow 20}}$ \\

\addlinespace[3pt]

\multirow{2}{*}{GPT-5.5} & OpenCode~\cite{cao2026agenticvbench} & $27.4 \pm 3.5$ & -- & 27 & 20 & 27 & 37 \\

& \textbf{OpenCode + VIVID} & $\mathbf{37.1 \pm 3.5}$ & \textbf{+9.7} & \textbf{35}$^{\textcolor{blue}{\uparrow 8}}$ & \textbf{41}$^{\textcolor{blue}{\uparrow 21}}$ & \textbf{27}$^{\textcolor{blue}{0}}$ & \textbf{46}$^{\textcolor{blue}{\uparrow 9}}$ \\
\bottomrule
\end{tabular}
}

\vspace{2pt}
\begin{minipage}{\textwidth}
\footnotesize
\end{minipage}

\end{table*}

As illustrated in Fig.~\ref{fig:method}(d), the agent can examine source materials according to the task brief and develop a candidate editing plan using video overviews, local views, or transcripts. During reasoning, it can retrieve earlier evidence, request additional observations, or record unresolved questions. It then implements the plan and renders an actual media file for output inspection.

Output inspection and revision connect the rendered result to subsequent editing decisions. The agent can review the output, return to relevant source evidence, revise its plan, and render again before delivering the task-required files. Observation, reasoning, editing, and review can be interleaved according to the agent's strategy. Persistent task state and integration with the native harness support this workflow.

\paragraph{Persistent task state.}
In addition to media evidence, VIVID preserves the original task brief and allows the agent to record its understanding of the materials, editing plan, and unresolved questions. Task notes retain this working state, while an automatically maintained index provides navigation to media versions and evidence. Together with the preserved workspace, they make prior requirements, decisions, and observations accessible when work resumes in a fresh context. Continuation requires reconciling the recorded state with current artifacts and unfinished processing jobs, so that subsequent decisions reflect the actual execution state. The cumulative task budget carries over.

\paragraph{VIVID with the native harness.}
VIVID extends the native harness while preserving the agent's choice of how to implement edits. The agent can use the native execution environment for specialized transformations and VIVID's optional editing support for explicitly declared operations such as trimming, concatenation, constant-speed changes, and caption placement. For declared edits, source--output time mappings connect positions on the declared output timeline to the corresponding source intervals, supporting source tracing during revision and the projection of source-time captions onto the output timeline. Supplementary material provides implementation details.

\section{Experiments}
\label{sec:experiments}

\subsection{Experimental Setup}
We evaluate VIVID on AgenticVBench, a benchmark comprising 100 video post-production tasks across four categories: Assembly, which requires selecting and assembling shots; Repair, which requires correcting timing or audiovisual defects; Sequencing, which requires restoring the order of clips; and Repurpose, which requires adapting footage according to editing instructions. We retain the benchmark's original task inputs and task-specific evaluation methods for the final videos.

The experiments cover three model--harness configurations: GPT-5.6-sol with Codex, Claude Fable 5 with Claude Code, and GPT-5.5 with OpenCode. Additionally, we evaluate GPT-5.6-sol in Crayotter~\cite{yan2026crayotter}, a harness specifically designed for long-horizon video editing tasks, to provide an additional same-model comparison. For the three primary configurations, VIVID extends each native harness with interfaces for media inspection, transcription, evidence retrieval, editing assistance, and output validation, while retaining the native shell and programming tools. VIVID does not train or update the underlying models, instead, it extends and improves the harness. Each task is run independently three times for each model--harness configuration under a fixed tool schema, with identical task inputs and matched rollout limits (Supplementary Material). We follow AgenticVBench's task-specific scoring protocols, including its chance-adjusted Assembly score. Runs that produce no usable output are assigned a score of 0.

\subsection{Main Results}

Table~\ref{tab:main_results} summarizes the main performance comparisons between native harnesses and their VIVID-integrated counterparts, alongside results for a selection of open-source and closed-source models and harnesses.

With VIVID, Codex, Claude Code, and OpenCode achieve overall average scores of \textbf{58.6}, \textbf{55.1}, and \textbf{37.1}, respectively, representing gains of \textbf{20.2}, \textbf{22.7}, and \textbf{9.7} percentage points over their respective native configurations. Scores improve across all task categories for Codex and Claude Code. For OpenCode, they improve on Repurpose, Sequencing, and Assembly and remain unchanged on Repair. These results demonstrate that VIVID can be integrated with diverse native harnesses and consistently improves overall performance across all three evaluated model--harness configurations. In an additional same-model comparison using GPT-5.6-sol, Codex + VIVID achieves a score of \textbf{58.6}, outperforming native Codex and Crayotter, which score \textbf{38.4} and \textbf{33.7}, respectively, further highlighting the performance advantages of the VIVID-integrated system across different harness environments.

Across task categories, Assembly achieves the highest average score within each of the three configurations, while the improvement on Repair is relatively limited. Comparing configurations, Claude Code achieves the highest average score of 59 on Sequencing, while Codex performs best on Repair and Repurpose, with average scores of 43 and 58, respectively. The model--harness configurations also differ in how they invoke and use the interfaces introduced by VIVID, and their performance gains vary. In Section~\ref{sec:analysis}, we further analyze agent behavior and interface usage to investigate how these differences relate to performance improvements.

\subsection{Ablation Study}
\label{sec:ablation}

To examine the contributions of the three core components introduced in Section~\ref{sec:method}, we compare full VIVID with three variants using GPT-5.6-sol with Codex. Each variant removes one core component:

\begin{itemize}
    \item \textbf{w/o on-demand audiovisual inspection:} Disables VIVID's task-directed inspection of source video and audio, including frame sampling, spatial-region inspection, and audio analysis. These observations support content selection and the localization of relevant source intervals.

    \item \textbf{w/o persistent evidence memory and retrieval:} Removes VIVID's persistent evidence records, memory index, and retrieval support, together with the files used to retain task requirements, editing plans, and working notes. These records support task continuation and evidence reuse across editing iterations.

    \item \textbf{w/o output inspection and revision:} Disables VIVID's audiovisual review and technical checks of rendered outputs. This removes the output feedback used to identify editing issues and guide further revision.
\end{itemize}

\begin{table}[t]
    \centering
    \caption{Ablation of VIVID's three core components using GPT-5.6-sol with Codex on AgenticVBench.}
    \label{tab:ablation}
    \setlength{\tabcolsep}{2pt}
    \resizebox{\columnwidth}{!}{
    \begin{tabular}{@{}lccccc@{}}
        \toprule
        \textbf{Configuration} & \textbf{Avg.} & \textbf{Repurpose} & \textbf{Sequencing} & \textbf{Repair} & \textbf{Assembly} \\
        \midrule

        Full VIVID & \textbf{58.6} & \textbf{57.7} & \textbf{57.9} & \textbf{43.0} & \textbf{77.1} \\

        \addlinespace
        w/o on-demand inspection & 42.6 & 46.0 & 39.9 & 36.0 & 46.8 \\

        \addlinespace
        w/o evidence memory & 47.8 & 54.4 & 47.4 & 22.9 & 59.9 \\

        \addlinespace
        w/o output review & 44.0 & 46.3 & 48.3 & 26.7 & 49.8 \\

        \bottomrule
    \end{tabular}
    }
\end{table}

\begin{figure*}[htpb]
  \centering
  \includegraphics[width=\textwidth]{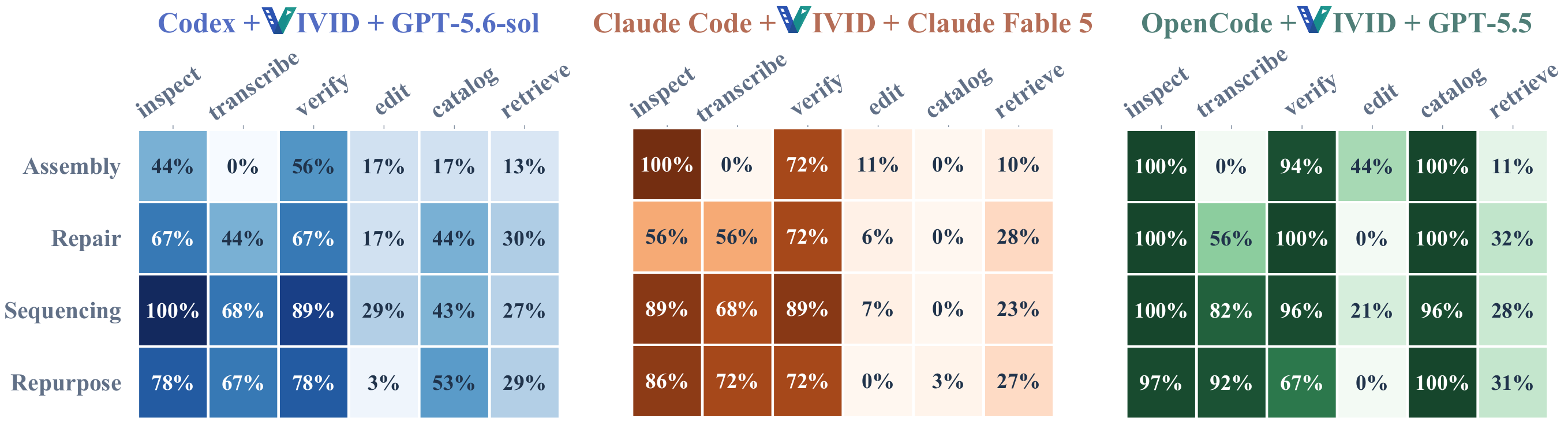}
  \caption{VIVID tool-use coverage across task categories and model--harness configurations. Each cell reports the percentage of task runs that invoke the corresponding tool at least once. Darker colors indicate higher coverage.}
  \label{fig:tool-call}
\end{figure*}

As shown in Table~\ref{tab:ablation}, full VIVID achieves an average score of 58.6. Removing on-demand audiovisual inspection reduces the score to 42.6, a decrease of 16.0 percentage points. Removing persistent evidence memory and retrieval reduces it to 47.8, a decrease of 10.8 percentage points. Removing output inspection and revision reduces it to 44.0, a decrease of 14.6 percentage points. Full VIVID outperforms all three variants across every task category.

On-demand audiovisual inspection produces the largest average performance drop when removed, with Assembly decreasing from 77.1 to 46.8. Removing persistent evidence memory and retrieval has its largest category-level effect on Repair, which decreases from 43.0 to 22.9. Removing output inspection and revision also substantially reduces Assembly and Repair scores, to 49.8 and 26.7, respectively. Together, these results support the contributions of acquiring audiovisual evidence, retaining evidence and task state across iterations, and using output feedback to guide revision.

\section{Model--Harness Behavior Analysis}
\label{sec:analysis}

\subsection{Tool Call Behavior}
\label{sec:tool-call}

Fig.~\ref{fig:tool-call} summarizes the task coverage of VIVID's six tools across the three model--harness configurations. We relate these usage patterns to the core components introduced in Section~\ref{sec:method}.

\begingroup
\makeatletter
\renewcommand\paragraph{\@startsection{paragraph}{4}{\z@}%
  {2ex \@plus .3ex \@minus .2ex}{-1em}{\normalfont\normalsize\bfseries}}
\makeatother
\paragraph{On-demand audiovisual inspection.}
The \texttt{inspect} tool is used in 44--100\% of tasks across the reported categories and configurations. Transcription coverage varies with task requirements: no configuration invokes \texttt{transcribe} for Assembly, while its coverage reaches 68--82\% for Sequencing and 67--92\% for Repurpose. This pattern is consistent with agents selecting visual and linguistic evidence according to the editing task, as supported by on-demand audiovisual inspection.

\paragraph{Persistent evidence memory and retrieval.}
The \texttt{retrieve} tool is used across all four task categories, with coverage of 10--13\% for Assembly, 28--32\% for Repair, 23--28\% for Sequencing, and 27--31\% for Repurpose. The similar coverage across configurations within each category shows that evidence retrieval is used across different native harnesses, supporting the role of persistent memory in making recorded observations accessible during editing.

\paragraph{Output inspection and revision.}
The \texttt{verify} interface is invoked in 56--100\% of tasks across categories and configurations. It provides file and decoding checks and output review planning, complementing audiovisual inspection through \texttt{inspect}. These calls show that agents use VIVID's technical checking support during task execution. Section~\ref{sec:case-study} further examines how inspection of rendered outputs informs subsequent revisions.

\endgroup
\subsection{Token Consumption and Performance Gains}
\label{sec:token-score}

\begin{figure}[t]
  \centering
  \includegraphics[width=\linewidth]{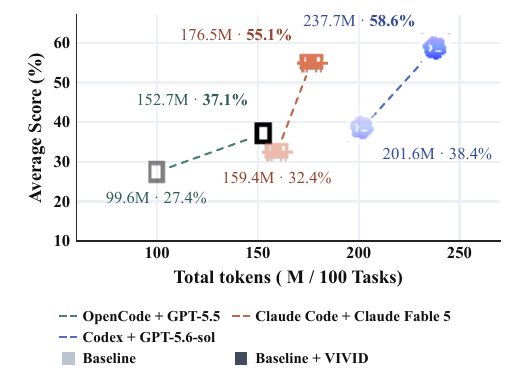}
  \caption{ Token consumption and editing performance with and without VIVID. The x-axis shows total token consumption per 100 tasks, and the y-axis shows the average task score. Dashed lines connect each native model--harness configuration to its VIVID-augmented counterpart. Lighter and darker markers indicate the native and VIVID-augmented configurations, respectively. }
  \label{fig:token-acc}
\end{figure}

To measure the inference overhead of VIVID, we compare token consumption and task scores for the three model--harness configurations with and without VIVID. Fig.~\ref{fig:token-acc} shows the total token consumption per 100 tasks and the corresponding average score. With VIVID, both token consumption and task scores increase across all three configurations.

Specifically, GPT-5.6-sol with Codex uses 17.9\% more tokens and improves its average score from 38.4 to 58.6, a gain of 20.2 percentage points. Claude Fable 5 with Claude Code uses 10.7\% more tokens and improves its score from 32.4 to 55.1, a gain of 22.7 percentage points. GPT-5.5 with OpenCode uses 53.3\% more tokens and improves its score from 27.4 to 37.1, a gain of 9.7 percentage points.

\subsection{Behavior Across Harnesses}

The Codex and Claude Code configurations make more selective use of VIVID's tools. As discussed in Section~\ref{sec:tool-call}, inspection and transcription have high coverage in Sequencing and Repurpose, while some other interfaces are used in fewer tasks. Both configurations also achieve larger performance gains with smaller increases in token consumption (Section~\ref{sec:token-score}). Together, these observations suggest more efficient integration of VIVID's tools with native capabilities, with agents drawing on media-specific support according to editing needs.

The OpenCode configuration makes broader use of some VIVID interfaces, particularly \texttt{catalog}, which is used in 98.9\% of its tasks. This pattern suggests a stronger tendency to adopt VIVID-provided interfaces across tasks. Its broader interface use is accompanied by a larger increase in token consumption and a smaller performance gain, yielding a less favorable balance between token overhead and score improvement in these experiments.

Across these different tool-selection patterns, VIVID improves the average task score for all three configurations: by 20.2 percentage points with Codex, 22.7 with Claude Code, and 9.7 with OpenCode. These consistent gains support the effectiveness of VIVID's design across the tested native harnesses. Its combination of on-demand audiovisual inspection, persistent evidence memory and retrieval, and output inspection and revision accommodates different agent workflows while improving editing performance.

\section{Trajectory Analysis}

\subsection{Analysis of Editing Processes and Outcomes}
\label{sec:editing-outcome-analysis}

\begin{figure*}[htpb]
  \centering
  \includegraphics[width=\textwidth]{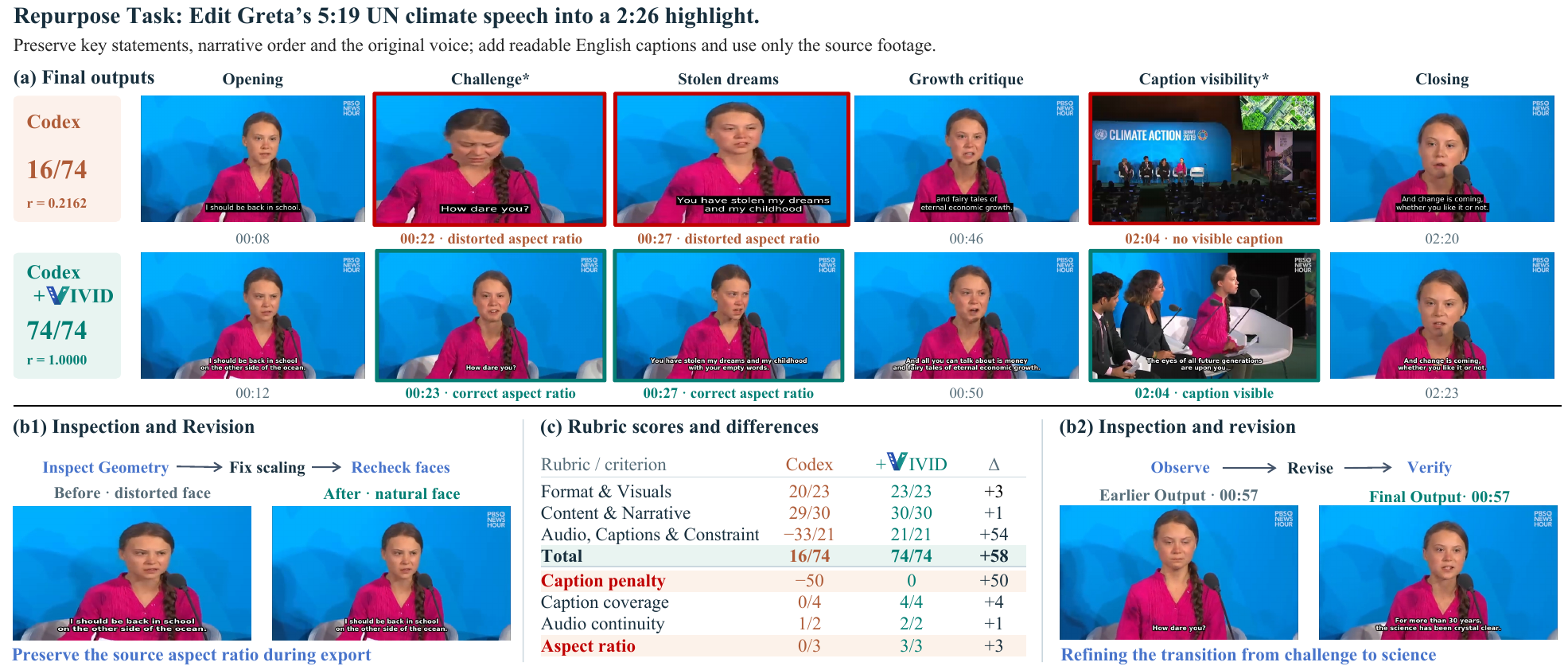}
  \caption{ Comparison of native Codex and Codex + VIVID. (a) Sampled frames from the final outputs show differences in aspect ratio and caption visibility. (b1--b2) Examples of inspection and revision with VIVID. (c) Rubric scores summarize the differences between the two outputs. }
  \label{fig:compare-case}
\end{figure*}

For each task, we randomly select one of three runs from native Codex and Codex + VIVID. Three hired video editing experts judge each trajectory against the applicable criteria for seven features: content selection, temporal localization, global ordering, visual composition, audiovisual alignment, repair quality, and plan--output consistency. Table~\ref{tab:editing-features} reports the percentage of applicable assessments that satisfy each criterion.

\paragraph{Content selection and temporal localization.}
With VIVID, the satisfaction rates for content selection (F1) and temporal localization and edit boundaries (F2) increase by 24.4 and 26.0 percentage points, respectively. Both require the agent to ground editing decisions in source observations: it must identify relevant content and locate the target events or defects. VIVID provides on-demand video inspection, local visual observations, and audio transcription, allowing the agent to obtain audiovisual evidence with temporal coordinates around candidate segments. These improvements are consistent with the intended role of inspection in supporting content selection and temporal localization. 

\begin{table}[t]
    \centering
    \caption{ Feature-level comparison of native Codex and Codex + VIVID. Each value reports the percentage of applicable evaluations satisfying the corresponding criterion. }
    \label{tab:editing-features}
    \small
    \begin{tabularx}{\columnwidth}{
        @{}
        l
        >{\raggedright\arraybackslash}X
        rr
        @{}
    }
        \toprule
        & \textbf{Feature} & \textbf{Codex} & \shortstack{\textbf{Codex + }} \\
        & & \textbf{(\%)} & \textbf{VIVID (\%)} \\
        \midrule

        F1 & Content selection & 32.6 & \textbf{57.0} \\

        F2 & Temporal localization & 43.0 & \textbf{69.0} \\

        F3 & Global ordering and narrative & 21.4 & \textbf{38.1} \\

        F4 & Composition and subject visibility & 45.8 & \textbf{62.5} \\

        F5 & Audiovisual alignment & 29.4 & \textbf{45.1} \\

        F6 & Repair and editing effects & 60.0 & \textbf{66.7} \\

        F7 & Plan--output consistency & 81.1 & \textbf{96.2} \\

        \bottomrule
    \end{tabularx}
\end{table}

\paragraph{Global narrative and audiovisual coordination.}
The satisfaction rates for global ordering and narrative coherence (F3) and for audio, captions, and audiovisual alignment (F5) increase by 16.7 and 15.7 percentage points, respectively. These requirements involve integrating information across clips or modalities: the former requires reasoning about ordering and narrative relationships, while the latter requires coordinating visuals, sound, and captions. VIVID's persistent evidence memory and retrieval component links audiovisual observations to media versions and temporal coordinates, allowing agents to revisit relevant evidence during editing. The same evidence can thus inform successive editing decisions. These gains are consistent with the component's role in supporting cross-clip and cross-modal judgments.

\paragraph{Visual composition and repair quality.}
The satisfaction rates for visual composition and subject visibility (F4) and for repair and editing effects (F6) increase by 16.7 and 6.7 percentage points, respectively. VIVID allows agents to inspect the visual and audio content of rendered outputs, providing feedback that can help identify problems such as subject cropping, poor framing, or insufficient repair. The improvement in F4 is consistent with the value of output inspection for visual composition. The smaller improvement in F6 also suggests that observation and verification need to be combined with effective processing methods, such as denoising, detail restoration, or color correction.

\paragraph{Consistency between the editing plan and the output.}
The satisfaction rate for plan--output consistency (F7) increases by 15.1 percentage points. This criterion evaluates whether the rendered output follows the editing plan. VIVID allows the agent to inspect rendered outputs, identify discrepancies with its plan, and revise the edit. For edits declared through VIVID, source--output time mappings link positions on the declared output timeline to the corresponding source intervals, supporting source tracing during output review and revision.

Overall, Codex + VIVID achieves higher satisfaction rates than native Codex across all seven features, with particularly large improvements in content selection and temporal localization. These results support VIVID's design of connecting source inspection, editing decisions, and output review through on-demand audiovisual evidence associated with media versions and temporal coordinates. At the same time, global narrative, audiovisual coordination, and repair quality remain areas for further improvement.

\subsection{Case Study: Speech Highlight Editing}
\label{sec:case-study}

To further examine how VIVID supports the editing process, we study a Repurpose task based on \emph{Greta Thunberg's UN climate speech}, as shown in Fig.~\ref{fig:compare-case}. The task requires the agent to condense a complete long-form speech into a concise highlight, preserving the key arguments, original narrative order, and speaker's voice while adding readable English captions and using only the source footage. The supplementary material presents additional case studies.

\paragraph{Final outputs.}
As shown in Fig.~\ref{fig:compare-case}(a), both configurations retain key moments from the speech, including the opening, the criticism of empty promises and economic growth, and the closing statement. They differ, however, in visual presentation and captions. In the native Codex output, some frames exhibit aspect-ratio distortion, and no caption is visible at one sampled time point. In the displayed Codex + VIVID frames, the speaker's proportions appear natural and the captions are visible. The evaluator assigns the two outputs scores of 0.21 and 1.00, respectively. Most of the difference comes from the audio, captions, and constraints category, in which the native output incurs a missing-caption penalty.

\paragraph{Inspection and revision.}
The Codex + VIVID trajectory shows that, after producing an initial render, the agent inspected an overview of the output and several specific time windows, then revised the captions and rendered the video again. During a subsequent review, the agent identified excessive repetition of the speech's signature challenge. It adjusted the edit and captions, using the recovered time to retain a complete statement about climate risk. After the revision, the agent inspected local windows again and checked the technical properties of the final file. This process illustrates how VIVID supports a render--inspect--revise workflow, allowing the agent to review specific intervals and make subsequent adjustments.

\section{Conclusion}

We presented VIVID, a visual harness that extends general-purpose coding agents to video editing. VIVID equips agents with on-demand audiovisual inspection, persistent evidence memory linked to media versions and temporal coordinates, and inspection of rendered outputs, enabling an iterative workflow of observation, reasoning, editing, and review while preserving their native tools and autonomy over editing decisions. Experiments on AgenticVBench show that VIVID consistently improves editing performance across diverse model--harness configurations without additional model training, and ablation studies confirm the effectiveness of each core component. These results demonstrate that grounding editing decisions in actively acquired, persistently retained, and revisitable audiovisual evidence is an effective path toward reliable and practical agentic video editing.

{
\small
\bibliographystyle{ieeenat_fullname}
\bibliography{bibliography/references}
}

\clearpage
\newpage
\appendix
\section{Method Details}
\label{app:method-details}

This section supplements Section~\ref{sec:method} with implementation details of VIVID's three core components: on-demand audiovisual inspection, persistent evidence memory and retrieval, and output inspection and revision. It then describes persistent task state and VIVID's integration with the native harness, including optional editing and declared time mapping, as supporting mechanisms for the iterative editing workflow.

VIVID exposes these capabilities through six default tools, defined in Table~\ref{tab:vivid-tools}. The tools support the core components and workflow alongside the native harness's file, shell, and code-execution tools. The agent retains control over what to inspect, how to implement edits, and when to review the output.

\begin{table}[htpb]
    \centering
    \caption{VIVID's six default tools and their definitions.}
    \label{tab:vivid-tools}
    \begin{tabular}{@{}p{0.26\columnwidth} p{0.64\columnwidth}@{}}
        \hline
        \textbf{Tool} & \textbf{Definition} \\
        \hline
        \texttt{catalog} & Browse source materials, registered media versions, and their media properties. \\

        \texttt{inspect} & Inspect video and audio on demand within specified time ranges or spatial regions, and compare multiple assets. \\

        \texttt{transcribe} & Transcribe a specified audio interval with segment-level or estimated word-level timing. \\

        \texttt{retrieve} & Search indexed text or revisit saved observations with their media-version and time references. \\

        \texttt{edit} & Optionally trim, concatenate, and change the speed of clips, process subtitles, compute declared mappings between source and output time, and manage native command jobs. \\

        \texttt{verify} & Check files, media properties, and decoding status, and plan output review windows based on a timeline or time-referenced requirements. \\
        \hline
    \end{tabular}
\end{table}

\subsection{Core Task Prompt}
\label{app:core-task-prompt}

VIVID prepends the following task prompt, lightly edited here for presentation, to the complete original task brief and the current persistent state, including the workspace, budget, continuation status, and recovery information. The prompt preserves the agent's choice of editing methods while distinguishing sampled observations and technical checks from evidence of content quality.

\begin{listingbox}[label={Core-Task-Prompt}]{Core Task Prompt}
\normalsize\normalfont
\setlength{\parindent}{0pt}
\setlength{\parskip}{0.5em}
Complete the original task using your native tools and any useful VIVID media tools.
The original task brief governs source permissions, constraints and deliverables.
Keep supplied sources unchanged.

VIVID's optional media helpers are described by the tool schemas. Use native tools
and choose your own editing methods; no plan registration or VIVID execution is required.

Sampled frames are not continuous coverage. A transcript or spectrogram is not
proof of listening. Technical checks do not establish content quality.

TASK.md preserves the brief. GUIDE.md and WORKING.md are optional native-file notes;
MEMORY.md indexes archived evidence.

When a fresh session is necessary, update notes
and end with VIVID\_CHECKPOINT. On continuation, inspect actual files and pending jobs.
Report delivered files and any unverified requirements.
\end{listingbox}

\subsection{On-demand Audiovisual Inspection}

\begin{figure*}[htpb]
  \centering
  \includegraphics[width=\textwidth]{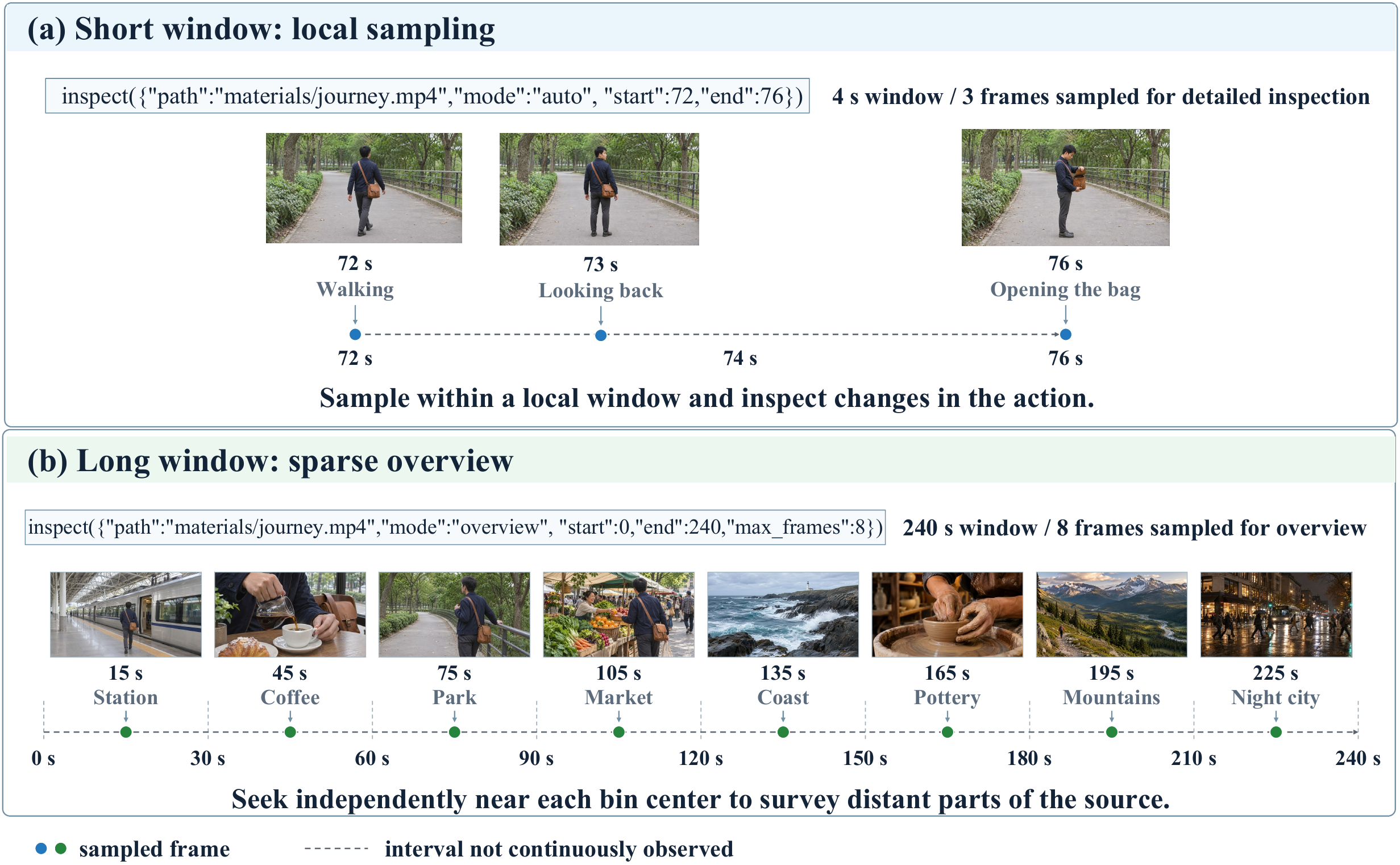}
  \caption{Video sampling in short and long windows. (a) A small number of frames are sampled within a short window to examine local changes in an action. (b) Frames are sampled across a long window to provide a sparse overview of different temporal locations.}
  \label{fig:short-long-window-sample}
\end{figure*}

VIVID does not require the agent to view an entire video at once and retain all relevant content in its context. Instead, it allows the agent to select what to observe as its questions evolve. The agent can request video frames from a temporal window, crop a spatial region, or compare locations across assets. Contact sheets retain asset identities and the timestamps of the decoded frames. For longer material, the agent can begin with a sparse overview and then inspect candidate intervals in greater detail. For a local question, it can sample directly within a shorter window, as shown in Fig.~\ref{fig:short-long-window-sample}.

The same principle applies to audio. The agent can inspect a selected audio window through amplitude statistics, waveforms, or spectrograms, and request transcription with segment-level or estimated word-level timestamps. Longer transcription requests may be processed in chunks, allowing completed intervals to be retained even when later intervals remain pending.

\subsection{Persistent Evidence Memory and Retrieval}

For an observation to remain useful throughout an editing task, the agent must be able to recover it and identify its origin. VIVID therefore stores acquired observations outside the current model context and associates them with evidence identifiers, media versions, temporal coordinates, and observation types. The current implementation identifies a registered version using the file's workspace-relative path and content hash.

This distinction is particularly important when a file is rewritten at the same path. Previously archived observations remain associated with the version from which they were obtained, while newly registered content receives its own version identity. The agent can therefore revisit an earlier observation without confusing it with the file's current contents.

The agent can retrieve evidence by identifier or search indexed transcripts, notes, and other records, narrowing results by asset, time range, or evidence type.

\subsection{Output Inspection and Revision}

VIVID reuses on-demand audiovisual inspection to examine the actual rendered output and combines these observations with technical checks to support revision. The verification interface checks file identity, media properties, decoding status, and specified format requirements. Given a declared timeline, subtitles, or task requirements with output-time coordinates, the system can also suggest review windows, including windows around cut points. 

The agent can use the observation interface to inspect the actual output and revisit source evidence as needed. If it finds a problem, it can revise the plan, render again, and inspect the new file.

\subsection{Persistent Task State}

VIVID uses files to preserve task requirements and working state for context continuation. \texttt{TASK.md} stores the original task requirements. The agent can record relatively stable interpretations of the material and editing constraints in \texttt{GUIDE.md}, and the current plan, unresolved issues, and next actions in \texttt{WORKING.md}. \texttt{MEMORY.md} is an automatically generated bounded index that provides navigation to media versions and recent evidence.

When a task needs to continue in a new context, the agent can resume work using the original requirements, notes, index, preserved workspace, and archived evidence, and revisit specific observations as needed. The cumulative task budget carries over rather than resetting at continuation. Notes represent a revisable working understanding and should be checked against the actual files, evidence, and job status when work resumes. For unfinished media-processing jobs, the agent should first inspect the actual process, outputs, and logs before deciding whether to continue waiting, cancel, or rerun the job.

\subsection{VIVID with the Native Harness}

VIVID preserves the agent's choice of how to implement edits. The agent can use the native harness's shell and code-execution tools with Python, FFmpeg, and other available programs, or use VIVID's optional editing support for explicitly declared trimming, concatenation, constant-speed changes, and caption operations. Specialized repair and visual effects can therefore continue to use the native execution environment.

\paragraph{Declared time mapping.}
VIVID's optional editing interface allows the agent to declare a sequence of source intervals and their constant playback speeds. From these declarations, the system computes the correspondence between the declared output timeline and the source intervals. Consider a sequence of concatenated segments. Let segment $i$ be taken from the source interval $[a_i,b_i)$, played at speed $r_i>0$, and placed at output start time $o_i$. In ideal continuous time, an output time $t$ within this segment maps to the source time
\begin{align}
    \tau_i(t) &= a_i+r_i(t-o_i), \\
    t &\in
    \left[o_i,\,
    o_i+\frac{b_i-a_i}{r_i}\right), \\
    o_i &= \sum_{j<i}\frac{b_j-a_j}{r_j}.
\end{align}
This correspondence allows the agent to trace a position on the declared output timeline back to the relevant source segment and project source-time captions onto the output timeline. It can support source tracing when revising an output.

\paragraph{Rendering and caption handling.}
The agent can render the declared clips and subtitles through VIVID or implement edits through the native execution environment. For subtitles crossing a cut boundary, VIVID's managed rendering uses available word-level timestamps to determine the retained text. Entries whose retained text cannot be determined are flagged for subtitle revision or additional timing information.

\section{Experiment Details}
\label{app:experiment-setup}

This appendix describes the experimental setup for evaluating VIVID on AgenticVBench, including the model--harness configurations, execution limits, and computational resources. Each model--harness configuration is evaluated in three independent runs per task. All experiments are conducted in Modal sandboxes.

\subsection{Evaluation Metrics}

We follow the task-specific evaluation protocols of AgenticVBench~\cite{cao2026agenticvbench}. Each task receives a normalized score in $[0,1]$, with higher scores indicating better performance. Runs that produce no usable output receive a score of zero. The four task families are evaluated as follows.

\paragraph{Assembly.}
The verifier checks whether the agent selects the correct source clip for each storyboard slot and adjusts the score for chance performance. Let $r$ denote the fraction of slots with correctly selected clips and $k$ the number of candidate clips per slot. The score is
\begin{equation}
    S_{\mathrm{assembly}} = \frac{r - 1/k}{1 - 1/k}.
\end{equation}
Random selection has an expected score of $0$, while selecting the correct clip for every slot yields a score of $1$.

\paragraph{Repair.}
A programmatic verifier measures repair quality by comparing the agent's output with the corrupted input and a reference output. For a measurement $M$ where a higher value is better, the normalized score is
\begin{equation}
    s = \operatorname{clip}\left(
        \frac{M_{\mathrm{out}} - M_{\mathrm{broken}}}
             {M_{\mathrm{golden}} - M_{\mathrm{broken}}},
        0, 1
    \right),
\end{equation}
where $M_{\mathrm{out}}$, $M_{\mathrm{broken}}$, and $M_{\mathrm{golden}}$ are the measurements for the agent's output, the corrupted input, and the reference output, respectively. The comparison direction is adjusted for measurements where a lower value is better.

For audio and visual defects located within a known time window, the score is
\begin{equation}
    S_{\mathrm{repair}}
    = 0.9\,s_{\mathrm{in}} + 0.1\,s_{\mathrm{out}},
\end{equation}
where $s_{\mathrm{in}}$ measures repair quality inside the window and $s_{\mathrm{out}}$ measures preservation of the uncorrupted content outside it. For timeline defects, the verifier compares the reported edit ranges with the ground truth and checks that the rendered output is consistent with the reported edits. The score is zero if the consistency checks fail.

\paragraph{Sequencing.}
The verifier compares the predicted clip order with the reference order using normalized distance (ND), longest increasing subsequence (LIS), and adjacent fidelity (ADJ). These metrics measure overall displacement, global relative order, and preservation of consecutive clip pairs, respectively. The final score is
\begin{equation}
    S_{\mathrm{seq}}
    = (1-\mathrm{ND}) \cdot \mathrm{LIS} \cdot \mathrm{ADJ}.
\end{equation}
The multiplicative scores mea alignment across all three dimensions. A perfect ordering receives a score of $1$.

\paragraph{Repurpose.}
Each output is evaluated against a task-specific set of binary rubric items. Format requirements are checked programmatically, while the remaining criteria assess visual presentation, narrative content, and sound. Applicable penalty items deduct points for critical violations of the brief. The resulting task score is normalized to $[0,1]$.

\subsection{Model--Harness Configurations}

We evaluate four model--harness configurations, as summarized in Table~\ref{tab:model-harness-configurations}. The command-line interface (CLI) version of each harness is fixed by its corresponding container image. Anthropic and OpenAI models are accessed through the OpenRouter API.

\begin{table}[htpb]
    \centering
    \caption{Harnesses, package or source versions, and corresponding models.}
    \label{tab:model-harness-configurations}
    \resizebox{\linewidth}{!}{%
        \begin{tabular}{llll}
            \hline
            Harness & Package / source & Version & Model \\
            \hline
            Codex CLI & \texttt{@openai/codex} & \texttt{0.166} & \texttt{gpt-5.6-sol} \\
            \multirow[t]{2}{*}{Claude Code} & \texttt{@anthropic-ai/} & \multirow[t]{2}{*}{\texttt{2.1.220}} & \multirow[t]{2}{*}{\texttt{claude-fable-5}} \\
                & \texttt{claude-code} & & \\
            OpenCode & \texttt{opencode-ai} & \texttt{1.2.27} & \texttt{gpt-5.5} \\
            Crayotter & GitHub release & \texttt{1.1.0} & \texttt{gpt-5.6-sol} \\
            \hline
        \end{tabular}%
    }
\end{table}

The pinned package versions determine the CLI environment and built-in defaults of each harness.

\subsection{Run Settings}

Table~\ref{tab:run-settings} summarizes the experimental settings applied consistently across all model--harness configurations and independent runs.

\begin{table}[htpb]
    \centering
    \caption{Benchmark run settings applied consistently across all model--harness configurations.}
    \label{tab:run-settings}
    \resizebox{\linewidth}{!}{%
        \begin{tabular}{lp{0.55\linewidth}}
            \hline
            Setting & Value \\
            \hline
            Maximum agent iterations & 200 \\
            Per-task wall-clock time limit & 1,800 seconds (30 minutes) \\
            Compute per Modal sandbox & 4 vCPUs, 8 GB RAM \\
            Model parameters & Provider defaults; no overrides applied to \texttt{temperature}, \texttt{top\_p}, \texttt{top\_k}, \texttt{seed}, or \texttt{max\_output\_tokens} \\
            Idle timeout per LLM call & Respective harness CLI default \\
            \hline
        \end{tabular}%
    }
\end{table}

\begin{figure*}[t]
  \centering
  \includegraphics[width=\textwidth]{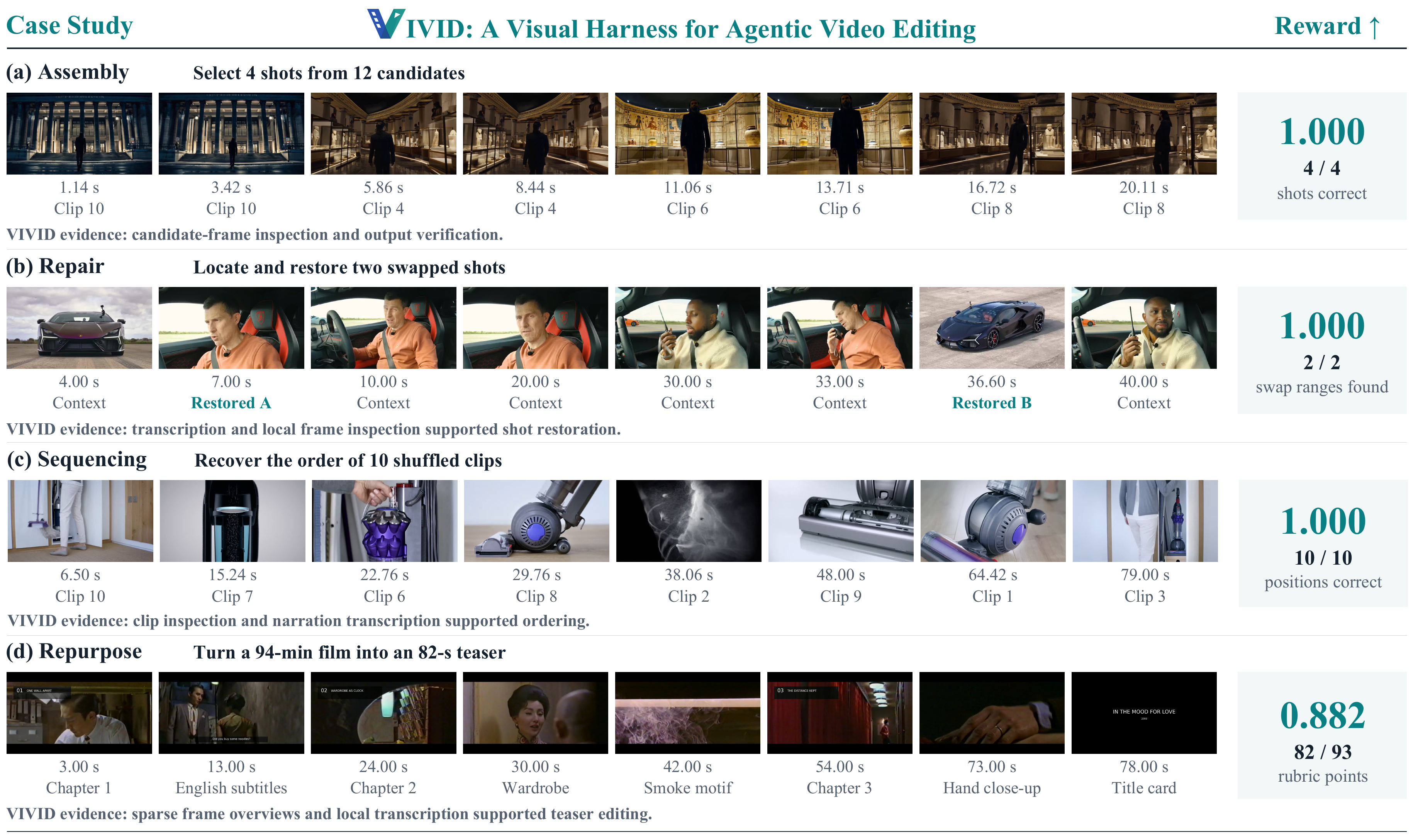}
  \caption{ Additional case studies across four editing tasks. (a) Assembly: selecting four shots from 12 candidates. (b) Repair: locating and restoring two swapped shots. (c) Sequencing: recovering the order of ten shuffled clips. (d) Repurpose: editing a long-form film into a short teaser. The examples illustrate how agents use VIVID's visual inspection and transcription interfaces to support editing decisions and output review. }
  \label{fig:additional-cases}
\end{figure*}

\subsection{Token Usage and Estimated Cost}
\label{app:token-usage-cost}

We access model APIs through OpenRouter\footnote{\url{https://openrouter.ai/}} and report token usage and estimated API costs for three model--harness configurations. Table~\ref{tab:token-prices} lists the token prices used for cost estimation. Table~\ref{tab:token-usage-cost} summarizes token usage and costs by task group. We report Repurpose separately from the other tasks because it uses Gemini 3.1 Pro and Claude Opus 4.7 as evaluators.

\begin{table}[htpb]
    \centering
    \caption{Token prices used for cost estimation, in US dollars per million tokens.}
    \label{tab:token-prices}
    \setlength{\tabcolsep}{4pt}
    \begin{tabular}{@{}lrrr@{}}
        \hline
        Model & Input & Cached input & Output \\
        \hline
        Claude Fable 5         & \$10.00 & \$1.00 & \$50.00 \\
        GPT-5.6-sol            & \$4.00  & \$0.40 & \$20.00 \\
        GPT-5.5                & \$5.00  & \$0.50 & \$30.00 \\
        Claude Opus 4.7        & \$5.00  & \$0.50 & \$25.00 \\
        Gemini 3.1 Pro & \$2.00  & \$0.20 & \$12.00 \\
        \hline
    \end{tabular}
\end{table}

\begin{table}[htpb]
    \centering
    \caption{Token usage and estimated cost by harness and task group. Token counts are in millions; costs are in US dollars. Per-run cost is the total cost divided by the number of runs.}
    \label{tab:token-usage-cost}
    \resizebox{\linewidth}{!}{%
        \begin{tabular}{llrrrr}
            \hline
            Harness & Task group & Input (M) & Output (M) & Cost (\$) & Cost/run (\$) \\
            \hline
            Codex & Non-Repurpose & 364.12 & 2.75 & 253.56 & 1.32 \\
            + GPT-5.6-sol    & Repurpose     & 344.25 & 1.92 & 217.47 & 2.01 \\
            Claude Code & Non-Repurpose & 274.21 & 3.64 & 607.32 & 3.16 \\
            + Claude Fable 5    & Repurpose     & 248.17 & 3.57 & 570.87 & 5.29 \\
            OpenCode & Non-Repurpose & 335.86 & 6.04 & 600.72 & 3.13 \\
            + GPT-5.5    & Repurpose     & 114.11 & 2.20 & 228.92 & 2.12 \\
            \hline
        \end{tabular}%
    }
\end{table}

\section{Additional Case Studies}
\label{sec:additional-trajectories}

As shown in Fig.~\ref{fig:additional-cases}, we further analyze representative trajectories from Assembly, Repair, Sequencing, and long-form video Repurpose tasks. These cases illustrate how VIVID supports source inspection, temporal localization, and output review across different editing scenarios.

\paragraph{Assembly and Sequencing.}
In Assembly task 1, Claude Fable 5 with Claude Code + VIVID actively inspects candidate frames to distinguish shot scale and camera movement. It selects four shots from 12 candidates, all of which match the reference. In Sequencing task 3, GPT-5.6-sol with Codex + VIVID combines visual observations across clips with speech transcription. It connects a shot of a vacuum cleaner being carried upstairs with a shot of the vacuum cleaner placed beside the same staircase, and all ten positions match the reference order. The agent also reports correcting an approximately 0.08-second AAC padding error and delivers an 86.04-second video. These cases illustrate how VIVID provides visual and linguistic evidence for shot selection and ordering, while supporting checks of the output file.

\paragraph{Repair.}
In the car shot-swapping task, Codex + VIVID uses discontinuities in dialogue, time-indexed transcription, and local frame inspection to locate the segments to be restored. The agent uses native commands to swap the input frame intervals $[145, 205)$ and $[882, 942)$, then calls VIVID's inspection tools to review the repaired output. The evaluator reports an IoU of 1.0 for both localized intervals and a mean frame MSE of zero relative to the reference. This case illustrates how VIVID helps the agent associate dialogue cues with visual content and temporal positions, supporting both repair localization and output review before delivery.

\paragraph{Long-form Video Repurposing.}
In the \emph{In the Mood for Love} teaser task, Codex + VIVID first generates a sparse 24-frame overview of the 5,640.04-second source film, then requests local word-level transcription for selected segments. It assembles an 82.17-second montage teaser and receives a rubric score of 82/93. This case illustrates VIVID's layered approach to long-form video: the agent first obtains a sparse visual overview, then acquires finer-grained evidence for candidate content and inspects the edited output.

\end{document}